# Machine Learning Framework for Early Power, Performance, and Area Estimation of RTL

Anindita Chattopadhyay
*Dept. of Electronics and Communication*
*BMS College of Engineering*
Bangalore, India
anindita.lvs21@bmsce.ac.in

Vijay Kumar Sutrakar
*Aeronautical Development Establishment*
*Defence Research and Development Organisation*
Bangalore, India
vks.ade@gov.in

*Abstract*— A critical stage in the evolving landscape of VLSI design is the design phase that is transformed into register-transfer level (RTL), which specifies system functionality through hardware description languages like Verilog. Generally, evaluating the quality of an RTL design demands full synthesis via electronic design automation (EDA) tool is time-consuming process that is not well-suited to rapid design iteration and optimization. Although recent breakthroughs in machine Learning (ML) have brought early prediction models, these methods usually do not provide robust and generalizable solutions with respect to a wide range of RTL designs. This paper proposes a pre-synthesis framework that makes early estimation of power, performance and area (PPA) metrics directly from the hardware description language (HDL) code making direct use of library files instead of toggle files. The proposed framework introduces a bit-level representation referred to as the simple operator graph (SOG), which uses single-bit operators to generate a generalized and flexible structure that closely mirrors the characteristics of post synthesis design. The proposed model bridges the RTL and post-synthesis design, which will help in precisely predicting key metrics. The proposed tree-based ML framework shows superior predictive performance PPA estimation. Validation is carried out on 147 distinct RTL designs. The proposed model with 147 different designs shows accuracy of 98%, 98%, and 90% for WNS, TNS and power, respectively, indicates significant accuracy improvements relative to state-of-the-art methods.*

## I. INTRODUCTION

In modern VLSI design flows, one of the most crucial stages wherein design engineers express the functionality of a design based on HDLs is RTL stage [1]. It allows significant flexibility to designers for performing experiment with different design alternatives for obtaining PPA of the target application specific integrated circuits (ASIC). Optimal RTL programming at this stage is highly critical because correcting the inefficiencies later in the synthesis phase is often very complex and, in most cases, infeasible [1]. More importantly, it is hard to assess the quality of RTL designs due to their representation as HDL code. Traditional VLSI workflows usually demand comprehensive synthesis or layout processes, which are based on advanced EDA tools in order to evaluate the quality of design based upon its post synthesis design or physical layouts. These processes take a long time, and for complex logic synthesis, more than a day is needed, while several days are taken by the layout stages. Further, the iterative nature of RTL optimization where synthesis tools are run repeatedly, results analyzed, and designs refined further prolong turnaround time and hampers overall efficiency [2, 3].

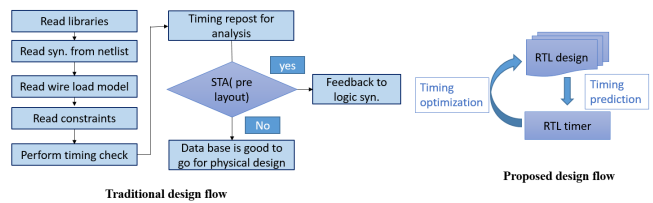

Fig 1. Traditional design flow versus proposed design flow of the RTL timing detection model.

Figure 1 shows the flow of conventional design versus the proposed methodology for RTL time detection model. The conventional approach makes use of final synthesis, wherein many months and computational resources are needed for estimating the time metrics. Several iterations of HDL code modifications have to be made by the designers, followed by synthesis invocation and netlists analysis at the gate level; hence the process is taking much time [4]. On the other hand, the proposed flow includes a streamlined RTL time detection model to eliminate the overhead of full synthesis altogether, transforming the HDL representation into a format suitable for machine learning enables predicting time parameters directly at the RTL level. Such an innovation results in remarkably reduced evaluation time, faster iterations, and an all-round improvement in efficiency without losing any accuracy in time analysis. The proposed framework represents a major step forward in early-stage design optimization within modern VLSI workflows.

Recent studies on ML explore ways to provide early-stage design feedback by predicting the quality of designs. Most of these methods focus on gate level netlist or layouts and leave RTL largely underexplored. Current ML techniques mostly use graphical neural networks (GNNs) in gate-level netlists or convolution neural networks (CNNs) in layouts [5, 6]. Since RTL designs are represented as HDL code rather than traditional data structures, there is no standardized method for representing and processing RTL. Some approaches adapt specific design flows without digging too much into RTL specifics, which ultimately demands retraining ML models for each new design. Current power and timing models at the RTL stage can neither scale well across a broad spread of designs nor scale well to combinational circuits generated by a specific set of RTL generators [7]. Recent advances in machine learning have resulted in more general approaches to predicting design attributes at RTL. Most such methods parse RTL designs into formats like abstract syntax trees (ASTs) and evaluate the PPA of design along specific paths or register trees derived from such representations [8]. Despite these advances, their abilities on novel, unseen RTL designs are still severely limited by several challenges. For instance, AST-based representations, being mainly adherent to all diversity in design kinds, would have their operations used to process these representations either induce redundancies or fail to capture important patterns. These difficulties call for better representations and modelling techniques [8, 9].

The proposed model introduces an RTL-stage PPA estimation framework that is far more accurate than previous ones for new RTL designs. This will allow for the cross-design of all significant PPA metrics-computation for WNS, TNS, power analysis and gate area. Two major challenges at the RTL-stage of PPA modelling are confronted through this framework: first is selecting an appropriate RTL representation for ML methods, it should resemble the post synthesis design. The proposed framework here uses the simple operator graph as a bit-level design representation using fundamental single-bit logic operations, promising better generalization than AST-like representations. Second, it concentrates on extracting meaningful patterns for accurate PPA estimation by using advanced techniques to capture critical design features from the SOG representation, guaranteeing precise predictions for each PPA objective [10, 11].

RTL design types and styles are diverse, making it essential to adopt representations that ensure high cross-design accuracy for various ML methods. Regarding the challenge of developing methodologies for distinct design objectives, such as power, performance, and area, a one-size-fits-all approach is insufficient. Since the underlying mechanisms behind these metrics vary significantly, customized estimation techniques are required for each objective. For instance, the time model introduced in this proposed framework is the first to explicitly map critical paths and associated delays between registers, leveraging the SOG representation in mapping registers to the post synthesis design. Similarly, the power model innovatively incorporates switching information as features, for power estimations [12]. The [12] uses a toggle file extracted from Synopsys prime time tool Switching Activity Interchange Format (.SAIF file). However, in the present work, 45nm library is used [13].

The proposed model also introduces a comprehensive PPA (power, performance, area) evaluation of RTL designs. Additionally, the framework employs tailored estimation techniques, uniquely capturing critical path details for time analysis and integrating switching data alongside module-level evaluation for power prediction, thereby offering more comprehensive and accurate insights into RTL design quality. Altogether, these innovations make for a rich solution to the problem of PPA modelling and accurate design optimization at RTL with desirable accuracy and efficiency in a VLSI design.

## II. METHODOLOGY

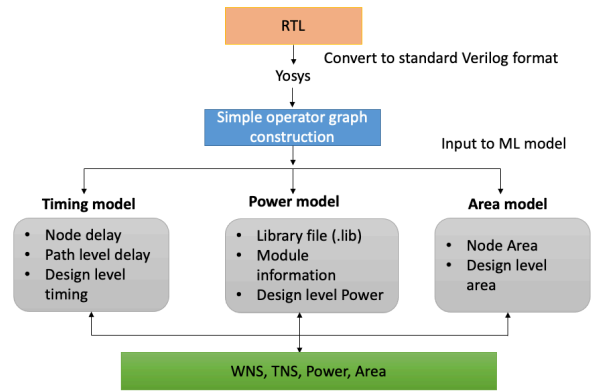

Fig 2. Overall workflow of the PPA prediction

This section provides a detailed explanation of the proposed model. Figure 2 depicts the comprehensive workflow of the proposed PPA prediction framework. The RTL is represented in HDL data and the corresponding post-synthesis is represented as $H$ and $G$. The time, power, and area characteristics represented as $\{P_G, T_G, A_G\}$. The goal of the proposed framework $F$, is to predict attributes for any given RTL design. The process begins by transforming the HDL-based representation $H$ into a format $R$, which facilitates detailed RTL analysis. Separate models for time, power, and area estimation, i.e. $\{f_t, f_p, f_a\}$ are then constructed [12]. The framework can be denoted as $F(H) = \{f_t(R), f_p(R), f_a(R)\} \rightarrow \{T_G, P_G, A_G\}$, where $R$ represented new RTL which is based on SOG architecture. Figure 3 illustrates the time evaluation process within the proposed framework, where all critical paths are identified and analyzed directly at the RTL stage. This approach enables early and accurate time predictions, streamlining the design process and improving overall efficiency in performance evaluation.

### A. SOG: Bit-Level RTL Representation

The proposed model at the RTL stage begins with transforming the HDL code ($H$) into structured format ($R$), to facilitate the intricacies of RTL design. One of the main challenges is bridging the significant gap between the RTL and post synthesis design, without relying on the computationally complex logic synthesis tool that requires significant time. Previous works have converted the HDL code into an AST-like representation [12 -14]. In AST- like structure, 18 word-level operations are possible. In contrast, the bit level approach eliminates the necessity for time consuming optimization steps in the conventional synthesis process, streamlining the workflow and reducing the overall processing time. This representation consists of single-bit registers and five basic logic operations AND, OR, XOR, NOT, and 2-to-1 multiplexer—forming SOG [12-15]. The bit level dataset required for SOG representation has been generated using open-source tools Yosys [16]. The key benefits of using the bit level representation of the RTL are it exhibits higher degree of similarity to the post- synthesis design, thereby reducing the disparity between the pre-synthesis RTL and the synthesised design, leading to more accurate predictions. Furthermore, the bit level employs just 5 basic single bit logic operations, and it is able to convey more complexity and versatility that the ATS structure, thereby reducing the variability among different RTL designs and allowing for a broader range of applications.

The proposed model supports these claims with experimental results shown in Table II. Apart from these advantages, SOG offers another benefit i.e. uniformity in register mapping, each register in the post synthesis design has a direct one-to-one mapping with single-bit registers in the SOG. This consistency allows for more precise and detailed time modelling, as critical paths in SOG can be mapped to equivalent paths in the post synthesis design. This feature significantly enhances time, power, and area models, which will be introduced in the following subsections.

*B. Time modelling*

The proposed framework for time evaluation presents a pre-synthesis framework for accurately predicting WNS and TNS using a combination of analytical modelling and machine learning techniques. The process begins with an RTL design, which is converted into a bit-level representation for detailed analysis. This step ensures the granularity required for precise delay and path analysis. It helps capture the data flow through individual bits in the design. Analytical node delays are computed for each element in the design, enabling the identification and mapping of critical paths by matching source and sink nodes. Critical paths, which are the longest delay paths in the circuit, are identified. This involves, (a) Source Matching (i,e. locating the starting nodes of each path) and (b) Sink Matching (i.e. Identifying the endpoints of the paths). These paths are further analyzed using a SOG to propagate delays and infer path models. The delays computed at the node level are propagated across the graph to determine the cumulative delays along each path and the model inference captures the overall timing behavior of each path, considering the relationships between the nodes. After this stage, a Random Forest machine learning model is trained using features extracted from the critical paths. The features include total number of operations in the path, the count of each type of operation (e.g., AND, OR, NOT) and the accumulated delay for the path, derived from the analytical delay model. This training enables the model to predict path-level time metrics effectively. The output from the Random Forest path model serves as input to an XGBoost ML model. This advanced ML algorithm refines the predictions by considering critical path characteristics. It predicts the TNS i.e. the sum of timing violations across all paths and WNS that is most severe timing violation in the design. The proposed model produces precise predictions for WNS and TNS. These results provide key insights into the timing performance of the RTL design at the pre-synthesis stage, enabling designers to address timing bottlenecks early in the design process. The detailed Time evaluation flow is shown in Figure 3 and discussed in the subsequent section.

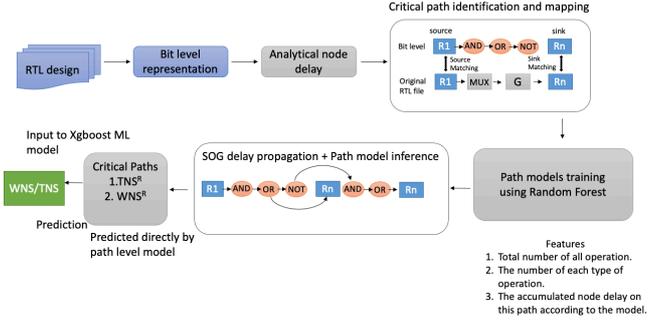

Fig 3. The proposed framework for time evaluation process that identifies all critical paths at the RTL, facilitating precise time predictions early in the design flow.

*1. Node delay modeling*

In the present work, a node delay model is taken from .lib file. In general, an analytical approach is used to extend the node delay modeling in *R*, to account for more complex interactions between fanout, wirelength, and capacitive effects. By integrating data from extracted parasitic, Resistance (R) and Capacitance (C) values and considering input transition times from the library file, the model can be refined to predict delays with greater accuracy across varying technology nodes.

*2. The identification and mapping of critical paths in R and G*

Critical path is identified by analyzing the RTL and SOG *R* and its corresponding graph *G* to locate the paths with the maximum delay between register pairs. This is done by identifying the start and endpoint registers, followed by tracing the dependencies through the respective graph representations. The goal here is to determine the longest combinational path in *R*, considering high-level constructs, and map it to its equivalent in *G*, which accounts for physical implementation details. The mapping ensures a direct correlation between the high-level design and gate-level implementation, enabling accurate time closure and time optimization in subsequent design stages.

*3. Path-level delay model*

For path delay prediction, the Random Forest model has been used which does the regression based on the given features using 80 estimators with a maximum depth of 20. This model's objective is mainly to estimate delays across different design routes. In the proposed model that path-level delay modeling focuses on accurately predicting the time of critical paths in a circuit. The training process involves extracting paths from the post-synthesis and associating them with ground-truth delays derived from timing reports. These paths are selected based on their significance, such as the top 1% maximum-delay paths, to ensure the model focuses on the most impactful time constraints. This approach provides a scalable solution to capture complex time interactions across varying design configurations.

*4. Path model inference*

The path model inference captures the delay at critical paths in a digital circuit, including the cumulative effects of individual node delays. In RTL-to-gate mapping, by analyzing paths $P^{G_{i \to j}}$, a model $f^{path}_t$ can make inferences about delays by aggregating node-level properties, such as fanout, capacitance, and cell delays. This inference process provides an overall perspective on time behaviour, including both combinational logic depth and sequential dependencies. Through the SOG, the proposed model propagates the expected node delays. Therefore, the model finds $N$ paths in total. Their estimates in the bit level representation $R$ are computed as $WNS^R$ and $TNS^R$.

*5. Estimation of WNS and TNS at design level*

In the proposed method, the direct path delay estimate $WNS^R$ and $TNS^R$, generated by path-level model serve as an initial approximation but lack the precision required for accurate design optimization. However, they offer helpful data for accurate estimation. The proposed model employs XGboost machine learning as final-stage model with the following features, i.e. (a) the design scale is indicated by the SOG characteristics, (b) the path-level model provides the estimated $WNS^R$ and $TNS^R$ and predicts the slack distribution of the worst 1% of the $N$ critical path. The XGBoost model uses aggregated data that have fine-tuned predictions that rely on 45 estimators having a maximum of depth 8. This stage ensures higher alignment between RTL predictions and gate-level time, enhancing the overall accuracy and reliability of early-stage timing analysis.

*C. RTL-Stage Power Modelling*

This module-level power modelling at the RTL stages can leverage switching activity as an input parameter. Unlike previous methods, this will annotate each node in the bit level representation with switching data such that it provides accurate power estimation during the design evaluation process. The switching activity data is, which represent dynamic transitions in the design, are obtained from the library file. The mapped information is stored onto the relevant nodes of the SOG representation and ensures preservation of functional relationships. Thus, the proposed methodology allows power estimation at precise levels without exhaustive simulation, enhancing the efficiency of RTL-stage power modelling [12,14,15].

The proposed model implements Graph Neural Network with a layered architecture, where two hidden convolution layers contain 10 and 70 nodes, respectively. A sum-pooling layer for the aggregation of graph-level features is also utilized. The GCN is trained to do end-to-end graph-level value regression optimized with the Adam optimizer with learning rate of 0.01. It converges very efficiently within 100 epochs, which demonstrates that it is efficient for module-level power estimation tasks. It uses power values when training a GCN, which inherently eliminates the dependence on explicit power simulation test benches. This approach can achieve accurate modelling of power by leveraging inherent features of design, thus useful in scenarios with limited access to detailed power simulation data.

*4. RTL-Stage Area Modelling*

The proposed model is much simpler than the other PPA models as discussed earlier in sections 2B and 2C. The area prediction model is designed, one-stage tree-based model can achieve high accuracy. Initially, the overall gate area is divided into (a) sequential and (b) combinational cell areas. The sequential area is estimated by multiplying the total number of registers in the SOG by the cell area of a D-flip-flop, as read off from the liberty file. This gives an accurate result, without further need for machine learning models. The combinational area is calculated by the area of all operators for the SOG using the liberty file and combines this data with the previously mentioned features of the SOG to construct a comprehensive feature set. Then, these features are applied in the tree-based model for the proper estimation of combinational area [12]. The area modelling approach combines both combinational area that is predicted by this model and the sequential area computed directly to get the total area.

There are three measures for verifying the model's accuracy is used in this paper. (a) The correlation coefficient (R), which assesses the degree of linear relationship between prediction and actual values; (b) the Mean Absolute Error Percentage (MAPE), which indicates the average percentage error; and (c) the Root Relative Square Error (RRSE), which normalizes the root-mean-square error against the standard deviation of the ground truth [13]. The dataset of n=147 designs is used for this assessment.

III. DATASET AND MODEL

The proposed framework is implemented by 147 distinct open-sources RTL designs from several benchmark sources (refer Table 1 for further details). All of the popular HDLs, such as Verilog, VHDL, Chisel, and Spinal HDL are used as initial codes. These codes are then converted into Verilog format. These designs are intended for a variety of applications, including as vector arithmetic, Advanced Encryption Standard (AES) core, Ethernet MAC core, Serial Peripheral Interface (SPI) core, cryptographic arithmetic, Hwacha: a vector processor generator, CPU cores, ML accelerators, and other designs for logic synthesis research. Open-source programs like Yosys [16] and Pyverilog [17] are used to build the bit-level SOG representation. The experiments are carried out on a workstation with NVIDIA RTX T400 graphics card and 32GB RAM 500GB ROM.

TABLE I. DATASET USED IN THE PROPOSED MODEL.

| Benchmarks | No. of designs | HDL type |
|---|---|---|
| ISCAS'89 [18] | 15 | Verilog |
| ITC'99 [19] | 22 | VHDL |
| OpenCores [20] | 48 | Verilog |
| VexRiscv [21] | 26 | Spinal HDL |
| RISC-V [22] | 10 | Verilog |
| NVDLA [23] | 8 | Verilog |

## IV. RESULTS AND DISCUSSIONS

Two different cases are considered with different benchmark circuits ISCAS'89 [18], ITC'99 [19], OpenCores [20], VexRiscv [21], RISC-V [22], NVDLA [23], Chipyard [24]. Case-I show the results of 90 benchmark circuits (non-optimized designs). Case-II shows a comparison of the reference model [12] with the proposed model (a) with 90 optimized designs using Yosys and (b) with additional benchmark circuits.

**Case I:** In Case I, the performance of the proposed framework was evaluated on 90 benchmark circuits without initial optimization. The predictions for SOG showed moderate accuracy, with a Pearson correlation coefficient of 0.83 for WNS and 0.9 for TNS. The MAPE values were 26% and 30% for WNS and TNS, respectively. Power predictions exhibited relatively higher error with a MAPE of 48% and an R value of 1, indicating challenges in estimating power consumption for unoptimized circuits. However, area predictions were more accurate, with an R value of 0.96 and a MAPE of 38%. These results suggest that while the framework demonstrates reasonable accuracy for unoptimized designs, improvements are necessary for predicting power and slack metrics.

**Case II:** In Case II, the evaluation was extended to 90 and 147 optimized benchmark circuits, where the framework demonstrated significantly enhanced accuracy across all metrics. For WNS, the correlation coefficient improved to 0.96 for the 90 optimized circuits and 0.98 for the 147 optimized circuits, with MAPE values reduced to 15% and 12%, respectively. TNS predictions achieved similarly high accuracy, with R values of 0.97 and 0.98 and MAPEs of 27% and 24%. Power and area predictions also improved substantially, with MAPEs dropping to as low as 33% and 12%, respectively, and RRSE values decreasing across the board. These results highlight the impact of circuit optimization on improving the framework's predictive performance, demonstrating its robustness and reliability for optimized designs.

The plots in figure 4 compare the predicted and measured values for four key performance metrics across multiple designs: (a) WNS, (b) TNS, (c) Power (in mW), and (d) Area (in mm²). Each plot illustrates a strong correlation between the predicted and ground-truth values, with data points closely aligning along the diagonal line representing ideal predictions. These results validate the effectiveness of the proposed methodology in accurately predicting performance, power, and area characteristics for all tested designs. Across all subplots, the proposed framework achieves consistently lower MAPE values compared to the reference, indicating significant improvements in prediction accuracy. These results establish the utility of the proposed approach in predicting PPA characteristics with high precision, ensuring better decision-making during the design process.

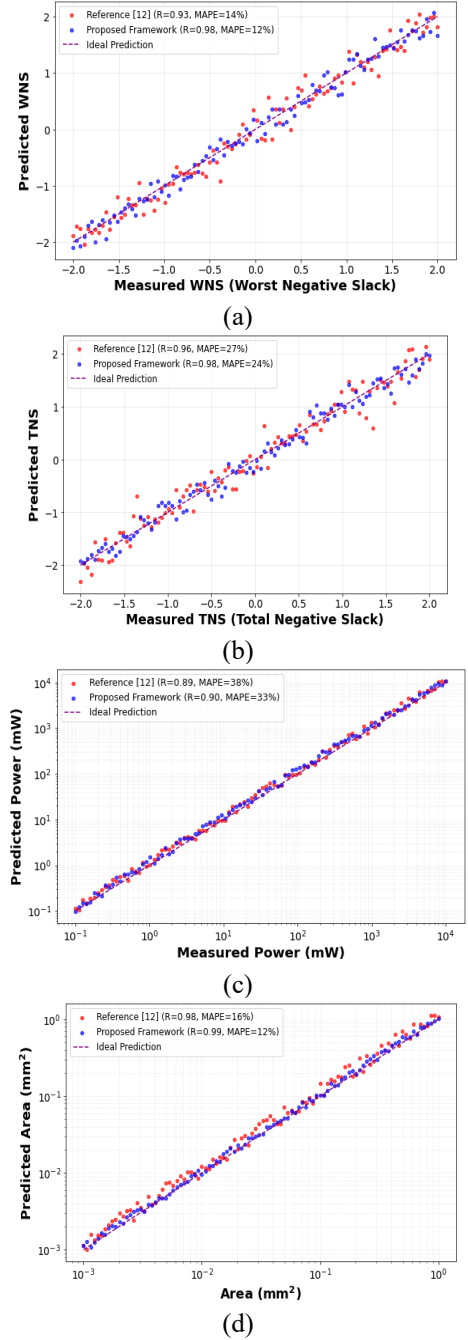

Fig 4. Prediction of (a) WNS, (b) TNS, (c) Power, and (d) Area of all designs.

## V. CONCLUSION

This research introduces a framework for predicting key PPA metrics, validated on 90 unoptimized and 90 and 147 optimized benchmark circuits. The framework shows moderate accuracy for unoptimized designs, but optimization significantly enhances its predictive performance, achieving strong correlations and reduced errors across all metrics. These results highlight the framework's reliability for early-stage design evaluation, especially in optimized scenarios. Future work will focus on improving predictive accuracy and scalability for more complex designs, contributing to efficient and reliable performance evaluation in IC design workflows.

TABLE I. ACCURACY COMPARISON FOR WNS, TNS, TOTAL POWER AND AREA EVALUATIONS (NO: NON-OPTIMIZED, O: OPTIMIZED; 90 AND 147 ARE THE DESIGNS CONSIDERED.

| Method | Target | R | MAPE | RRSE | Target | R | MAPE | RRSE | Target | R | MAPE | RRSE | Target | R | MAPE | RRSE |
|---|---|---|---|---|---|---|---|---|---|---|---|---|---|---|---|---|
| AST_90 [12] | WNS | 0.81 | 22% | 0.6 | TNS | 0.95 | 36% | 0.31 | Power | 0.79 | 44% | 0.63 | Area | 0.94 | 31% | 0.34 |
| SOG_90 [12] | WNS | 0.93 | 14% | 0.4 | TNS | 0.96 | 27% | 0.29 | Power | 0.89 | 38% | 0.54 | Area | 0.98 | 16% | 0.24 |
| AST_90 NO | WNS | 0.77 | 37% | 0.9 | TNS | 0.98 | 38% | 1.4 | Power | 0.78 | 53% | 1 | Area | 0.96 | 42% | 0.6 |
| SOG_90 NO | WNS | 0.83 | 26% | 0.6 | TNS | 0.9 | 30% | 0.39 | Power | 1 | 48% | 0.7 | Area | 0.96 | 38% | 0.4 |
| AST_90_O | WNS | 0.82 | 22% | 0.7 | TNS | 1 | 38% | 0.32 | Power | 0.78 | 38% | 0.5 | Area | 0.94 | 27% | 0.3 |
| SOG_90_O | WNS | 0.96 | 15% | 0.4 | TNS | 0.97 | 27% | 0.27 | Power | 0.92 | 27% | 0.2 | Area | 0.98 | 14% | 0.11 |
|  | WNS |  |  |  | TNS |  |  |  | Power |  |  |  | Area |  |  |  |
| AST_147_O | WNS | 0.83 | 19% | 0.4 | TNS | 0.88 | 36% | 0.3 | Power | 0.82 | 39% | 0.46 | Area | 0.96 | 26% | 0.28 |
| SOG_147_O | WNS | **0.98** | **12%** | **0.3** | TNS | **0.98** | **24%** | **0.24** | Power | 0.9 | 33% | 0.39 | Area | **0.99** | **12%** | **0.19** |